\documentclass{Interspeech2024}

\interspeechcameraready

\title{Speak \& Improve Corpus 2025: an L2 English Speech Corpus for Language Assessment and Feedback}

\name[affiliation={1,3}]{Kate}{Knill$^\ast$}
\name[affiliation={2}]{Diane}{Nicholls$^\ast$}
\name[affiliation={1,3}]{Mark J.F.}{Gales}
\name[affiliation={1}]{Mengjie}{Qian}
\name[affiliation={2}]{Pawel}{Stroinski}

\address{
  $^1$ALTA Institute/MIL Lab, Dept. of Engineering, University of Cambridge, UK\\
  $^2$Cambridge University Press and Assessment, UK\\
  $^3$Enhanced Speech Technology Ltd, UK
  }
\email{kmk1001@cam.ac.uk, diane.nicholls@cambridge.org}


\begin{document}

\maketitle

\begin{abstract}

    We introduce the {\it Speak \& Improve Corpus 2025}, a dataset of L2 learner English data with holistic scores and language error annotation, collected from open (spontaneous) speaking tests on the \href{https://speakandimprove.com}{Speak \& Improve} learning platform.
    The aim of the corpus release is to address a major challenge to developing L2 spoken language processing systems, the lack of publicly available data with high-quality annotations. 
    It is being  made available for non-commercial use on the \href{https://englishlanguageitutoring.com/datasets/speak-and-improve-corpus-2025}{ELiT website}.
    In designing this corpus we have sought to make it cover a wide-range of speaker attributes, from their L1 to their speaking ability, as well as providing manual annotations. This enables a range of language-learning tasks to be examined, such as assessing speaking proficiency or providing feedback on grammatical errors in a learner's speech. Additionally the data supports research into the underlying technology required for these tasks including automatic speech recognition (ASR) of low resource L2 learner English, disfluency detection or spoken grammatical error correction (GEC).   
    The corpus consists of around 315 hours of L2 English learners audio with holistic scores, and a subset of audio annotated with transcriptions and error labels.  
\end{abstract}

\section{Introduction}
% significance of releasing this dataset, refer to the IS2025 proposal
Spoken language assessment and feedback are crucial components of language learning. Developing robust and inclusive automated tools for these tasks remains a significant challenge. One important limitation to developing systems is the lack of high-quality labelled data. To address this issue and encourage research in these areas, we are distributing a new annotated corpus of spoken L2 (second language) learner English, collected from speaking tests performed on the \href{https://speakandimprove.com}{Speak \& Improve (S\&I)} learning platform~\cite{nicholls23_interspeech}. The \href{https://englishlanguageitutoring.com/datasets/speak-and-improve-corpus-2025} {\it Speak \& Improve Corpus 2025} consists of around 315 hours of recordings of L2 English learner speech on open speaking tasks, annotated with English speaking proficiency scores. In addition a subset of the data, around 55 hours, was  manually transcribed, including disfluencies, and grammatically error corrected phrases~\cite{knill23_slate}. The S\&I platform was used by speakers from around the globe yielding a wide range of L1 (first language) backgrounds. The proficiency levels of the speakers range from Elementary (A2) to Advanced (C1) on the CEFR scale~\cite{council2001common}. This breadth of speaker attributes and abilities is crucial for the development of inclusive language learning technologies.   We believe this is the most comprehensive L2 English learner public corpus released to date. 
%This will provide researchers with public access to L2 English learner speaking resources that have not been available in such a comprehensive format to date. 

%The inclusion of learners from a broad range of proficiency levels and linguistic backgrounds is crucial for the development of inclusive language learning technologies. 
Data in the corpus was collected from practice tests done on the S\&I web application. Learners take a multi-level, monologic (computer prompt:human response), English speaking practice test. This is similar in format to exams such as \href{https://www.cambridgeenglish.org/exams-and-tests/linguaskill/}{Cambridge University Press \& Assessment's 
Linguaskill Speaking Test},  
\href{https://www.pearsonpte.com}
{Pearson's PTE} and  \href{https://englishtest.duolingo.com/test_takers}{Duolingo's English Test}. Learners undertake a series of open speaking tasks\footnote{The S\&I speaking test includes a read aloud component but this is not included in this corpus.} each of which aims to examine a different aspect of their speaking ability. Scores are given for each task, or part, and over the complete test. The latter takes into account the candidates skills at use of  language resource, coherence, hesitation/extent and task achievement as well as their pronunciation/intelligibility. S\&I's open speaking nature means that the learners are better able to demonstrate their competency in spoken English than is the case for read speech tests. This does mean, however, that the text of what the learner said is unknown. 
Additionally, their speech is relatively spontaneous so includes natural speaking effects such as disfluencies (filler words, false starts, repetitions) and not necessarily grammatically correct sentences from a writing perspective. This is in addition to grammatical errors that they might make related to their level of proficiency in English. The range of proficiency levels and L1s will cause a wide variety of pronunciation errors relating to the learner's L1 accent and whether they know how to pronounce an English word.

The S\&I Corpus 2025 supports research on a wide range of language learning and assessment tasks including Spoken Language Assessment (SLA), Spoken Grammatical Error Correction (SGEC), and Spoken Grammatical Error Correction Feedback (SGECF). This will be explored by participants in the \href{https://sites.google.com/view/slate-2025/home#h.p45mnlhypot5}{Speak \& Improve Challenge 2025}~\cite{2024sandi_pipeline} who will have access to a pre-release of the corpus December 2024-March 2025, prior to the full public release of the corpus in April 2025.

Related work, the Speak \& Improve application, annotation of the data and the corpus selection and distribution are described in following sections.

\section{Related Work}

While there are other publicly available datasets, none match the breadth and diversity of the Speak \& Improve dataset for L2 English assessment and feedback. For instance, the International Corpus Network of Asian Learners of English (ICNALE)~\cite{ishikawa2023icnale} focuses exclusively on Asian L1s and only features CEFR levels from A2 to B2. The latter are pre-assigned based on the learners' previously obtained language certificates rather than based on their actual speech. Similarly, the CLES corpus~\cite{coulange2024corpus} only features L1 French speakers whose CEFR ranges from B1 to B2. The LearnerVoice dataset~\cite{kim2024learnervoice} presented at Interspeech 2024 contains audio and annotations on grammatical errors and disfluencies, but is not publicly available yet, although its authors are planning to release it. It is restricted to L1 Korean speakers.  Other L2 corpora such as speechocean762~\cite{zhang2021speechocean} and L2-ARCTIC~\cite{zhao2018l2arctic}, only focus on pronunciation assessment of read speech. Existing English datasets for automatic speech recognition (ASR) training and evaluation of L2 learner speech, such as the ETLT Italian-L1 dataset~\cite{gretter2021etlt}, also tend to focus on single L1 groups, limiting their generalisability. The version of the ETLT corpus with annotated analytic and holistic scores~\cite{gretter2020tlt} has not been made public.

No current public dataset provides comprehensive support for spoken language assessment or spoken grammatical error correction/feedback at the scale of Speak \& Improve. Datasets such as the NICT-JLE~\cite{izumi2004nictjle} and KISTEC~\cite{kanzawa2022kistec} support disfluency detection, but only the former has annotations on grammatical error corrections. In both cases, the only available data consists of manual text transcriptions. The respective audio recordings are not available. 

While written GEC is an established area of study with five shared tasks in the last 15 years (i.e., the HOO 2011 Pilot Shared Task~\cite{dale2011hoo}, the CoNLL-2013 Shared Task~\cite{ng2013conll}, the CoNLL2014 Shared Task~\cite{ng2014conll}, the BEA-2019 Shared Task~\cite{bryant2019bea}, and MULTIGEC-2025~\cite{multigec2024}), one of the key innovations of this corpus is that it is the first to provide audio with grammatical error corrections to enable more complete research on spoken GEC. The complexities of spoken GEC, such as handling disfluencies, varied accents, and spontaneous speech patterns, make this task significantly different from written GEC, requiring new approaches and innovation. 

The Speak \& Improve Challenge 2025 demonstrates how the S\&I Corpus can be used for spoken language assessment and spoken grammatical error correction feedback, as well as for producing underlying technology such as automatic speech recognition and disfluency detection and grammatical error correction. For written assessment, the Automated Student Assessment Prize (ASAP) competition for essay scoring was organised in 2012 by the Hewlett Foundation~\cite{hammer2012asap}, and focused on L1 learners.   There have been other spoken language assessment shared tasks but they have generally targeted specific aspects such as pronunciation, grammar, or semantic meaning~\cite{baur2017overview,baur2018overview,baur2019overview}. 

\section{Speak \& Improve}
Speak \& Improve (S\&I) is a research project from the University of Cambridge in association with Cambridge University Press \& Assessment (CUP\&A) and English Language iTutoring Ltd (ELiT). The S\&I web application has been developed by ELiT with automated speaking assessment technology provided by Enhanced Speech Technology Ltd, developed through technology transfer from the ALTA Institute~\cite{nicholls23_interspeech}. 
Always available and free, users can interact with S\&I through many different devices including laptops, tablets and mobile phones, as it is based on the browser.
Learners are able to practice their English speaking and improve their confidence on a wide range of communicative speaking tasks. S\&I is designed for all proficiency levels, from basic beginner through independent intermediate to proficient learners; on the internationally-recognised CEFR~\cite{council2001common} scale from below A1 to C1 and above. 

The S\&I Corpus 2025 is selected from data collected by Speak \& Improve version 1. There were 1.7 million users of this version from across the globe from its launch in December 2018 to its retirement in September 2024. Over 18.4 million answers were submitted in total during that time. Learners were offered one of five complete practice tests to perform. They could choose to do the whole test, which takes around 10-15 minutes, or to stop after one or more parts of the test. The test structure is based on the Linguaskill Speaking test~\cite{xu2020linguaskill} which is designed to cover a variety of communicative speaking functions~\cite{xu2020linguaskill}.  A single test consists of 5 parts, always presented to a user in the same order:
\begin{itemize}
    \item {\bf Part 1: Interview}. \\
    The learner is asked 8 questions about themselves e.g. "Who in your family are you most similar to?". They are given 10 seconds of speaking time for the first 4 questions, and 20 for the second 4. The first two questions which may contain personal identity information are not marked and are not included in this corpus.
    \item {\bf Part 2: Read Aloud}. \\The learner must read aloud 8 sentences. 
    \item {\bf Part 3: Long Turn 1: Give your opinion}. \\The learner has 1 minute to give their opinion on a specific topic using 3 questions to guide them. 
    \item{\bf Part 4: Long Turn 2: Give a presentation about a graphic}. \\The learner has 1 minute to describe a process depicted in a diagram. 
    \item {\bf Part 5: Communication Activity: Answer questions about a topic}.\\ The learner has to respond to 5 questions relating to an overall topic, each for up to 20 seconds. 
\end{itemize}
At the end of each part the user receives an auto-marker assessment in the form of a CEFR-like level, along with an estimate of how confident the auto-marker is that the predicted score is a true reflection of the user's speaking level. Users who complete the test are awarded an indicative CEFR grade for the whole test. Unlike the Linguaskill test where restrictions are put on the length of thinking time test takers have before responding, the user can take as much time as they would like to prepare before answering each question. The questions for the tests were written exclusively for S\&I and do not form part of any CUP\&A test. 

Since the test takes some time many users drop out before the completion of the full test. Whilst only complete recordings are used for the test sets in the S\&I Corpus, some recordings from partially completed tests are included in the corpus as part of the data for training spoken language assessment systems.
Version 2 of S\&I has addressed this issue by offering users the chance to practise a variety of specific speaking skills in addition to doing complete tests.

\section{Corpus}

This section describes the S\&I Corpus 2025 including the data annotation, selection and distribution.
The first release of the corpus is split into three data sets: Train, Development (Dev) and Evaluation (Eval). These sets are being made available as a pre-release for the Speak \& Improve Challenge 2025~\cite{2024sandi_pipeline}.

\subsection{Data Annotation}
\begin{figure}[th]
    \centering
    \includegraphics[width=0.4\textwidth]{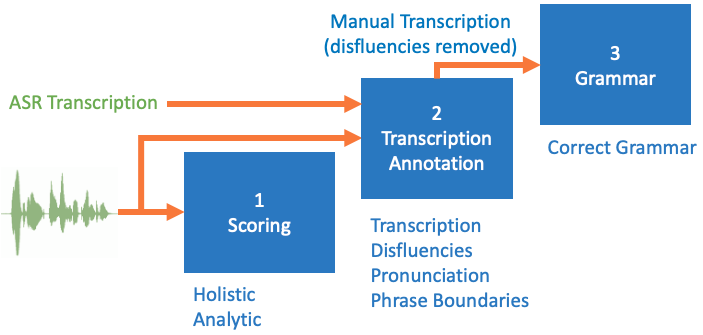}
    \caption{Annotation phases.}
    \label{fig:ann_phases}
\end{figure}

The S\&I data was annotated by ELiT human annotators using their bespoke annotation tool. They use a three stage approach as shown in Figure~\ref{fig:ann_phases}. This section describes the key points of the annotation with respect to the corpus. Further details can be found in~\cite{knill23_slate}.

\textbf{Phase 1: Scoring}. Utterances are initially given an audio quality score (range 3-5, utterances with an audio quality less than 3 were removed). Each test part, comprising one or more utterances, is then awarded an holistic score (scoring range 1-6). This holistic score summarises the annotator's overall/aggregated impression of the speaker's performance across all the utterances in the given part.  Parts receiving a holistic score equivalent to CEFR level A2 or higher are considered of sufficient quality and passed to the next annotation phase. Parts scored as A1 or below, in addition to those with poor audio quality, lack of meaningful responses, or inappropriate content (e.g. foul language), are excluded from further annotation and from consideration for the S\& I Corpus 2025.

\textbf{Phase 2: Transcription Annotation}. The second annotation phase generates the manual transcriptions of the audio. For efficiency reasons, these transcriptions were derived by correcting the output of an Automatic Speech Recognition (ASR) system\footnote{The ASR system used at this stage is not related to the Whisper models used in the distributed baseline pipeline for the Speak \& Improve Challenge 2025.}. The goal of this stage is to produce a detailed transcription that accurately reflects exactly what the learner said - including all language errors, hesitations, false starts and repairs and code-switching. In addition annotators mark phrase boundaries when they occur within an utterance. 
Word-level pronunciation errors are also indicated\footnote{For these pronunciation errors annotators were asked to ignore accent errors, and focus on lexical errors. For more details see~\cite{knill23_slate}.}. Where annotators were unable to identify a word, or words, in the audio these are tagged as unknown words. Where it is clear that the user is saying a proper noun which is foreign to British English these words were tagged as such to avoid the annotator having to guess what the user had said. Similarly code-words, where the user switched into speaking another language, were tagged. 
The associated phrases containing unknown, foreign proper nouns and/or code-switched words are excluded from the final annotation stage\footnote{These phrases are also excluded from ASR evaluation in the Speak \& Improve Challenge 2025.}. Tables~\ref{tab:annotation-marks}, ~\ref{tab:ann-word-tags} and ~\ref{tab:ann-phrase-tags} show the marks and tags that annotators can apply in phase 2 and are provided with the corpus transcriptions.

\begin{table}[htbp]
    \centering
    \begin{tabular}{l|l}
    \toprule
   Mark &  Description \\
   \midrule
    backchannel & Speaker spoke to themselves \\
    disfluency & Word is part of a false start or repeated disfluent sequence \\
    partial & Only part of the word was spoken \\
    pronunciation & Lexical pronunciation error made\\
    \bottomrule
    \end{tabular}
    \caption{Phase 2 annotation marks attached to a transcribed word.}
    \label{tab:annotation-marks}
\end{table}

\begin{table}[htbp]
    \centering
    \begin{tabular}{l|l}
    \toprule
      Tag   &  Description \\
      \midrule
          hesitation & Speaker made a sound indicating a hesitation e.g. um, er \\
          %\hline
      code-switch & Word is code-switched into language other than English \\
    foreign-proper-noun  & Foreign proper noun \\

    unknown & Annotator was unable to determine what the word spoken was \\
\bottomrule
    \end{tabular}
    \caption{Phase 2 word level transcription tags. The word in the transcript matches the tag i.e. is not a transliteration of what was said. Phase 3 annotators may tag a word as unknown if they are unsure how to correct the word.}
    \label{tab:ann-word-tags}
\end{table}
    
\begin{table}[htbp]
    \centering
    \begin{tabular}{l|l}
    \toprule
      Tag   &  Description \\
      \midrule
    speech-unit-incomplete & Partial phrase/speech unit\\
    speech-unit-statement & Phrase boundary ending with a full stop (period) \\
    speech-unit-question & Phrase boundary ending with a question \\
    \bottomrule
    \end{tabular}
    \caption{Phase 2 phrase level transcription tags. Phase 3 annotators may adjust phrase boundaries but these utterances were excluded from the corpus.}
    \label{tab:ann-phrase-tags}
\end{table}

\textbf{Phase 3: Error Annotation}. The final annotation phase focuses on correcting learner language errors. Phase 3 annotators are provided with texts only, to avoid any conflict of interpretation or second guessing by them as to whether the phase 2 annotator was correct. This phase aims to  generate an accurate transcription of the learner's intended speech. Prior to starting annotation, fluent phrase-level transcriptions are made from the disfluent utterance-level transcriptions output by phase 2. This is done by first splitting the utterance into phrases using the phase 2 phrase tags. Since it will not be possible to say what the correct grammar should be if only a partial phrase was spoken and/or it contains words that the annotators were unable to transcribe at phase 2, phrases containing any of these are excluded from error annotation at this point; i.e. phrases tagged as incomplete, or including words tagged as unknown, foreign proper nouns and/or code-switched.
Fluent transcriptions of the remaining phrases are then created by removing hesitations, partial words, false starts and repetitions from the manual disfluent transcription; these are marked as a disfluency or partial at phase 2. For example, {\em ``i am sti- still care about \%hesitation\% the the environment"} will become {\em ``i am still care about the environment"}. 
Using this fluent transcription the annotators generated the grammatically corrected transcription, e.g. {\em ``i still care about the environment"}. By comparing the fluent transcriptions to these corrected transcriptions it is possible to obtain a set of reference edits that can be fed back to learners.
Phase 3 annotators can mark a word with an error tag where the annotator knows the phase 2 annotator overlooked something or created a typo. They may also amend the phrase boundaries. Utterances with these edits were excluded from inclusion in the corpus for ease of processing.

\subsection{Data distribution and selection}

The data for the S\&I Corpus 2025 was selected from tests recorded on version 1 of the S\&I platform from 2019 to 2024. Learners were guided to do a full practice test. All data underwent Phase 1 holistic scoring at the test part level. A subset of the data received manual transcription annotation at Phase 2. The fluent phrases from Phase 2 were then passed to Phase 3 for grammatical error annotation. Annotators at phase 1 identified responses with poor audio recording quality and/or aberrant/malpractice responses. These were excluded from consideration for the corpus.

The Dev and Eval test sets were first defined to consist of full test submissions only i.e. where the learner has completed the test and all parts were successfully annotated at each phase of interest.
300 fully annotated test submissions were selected for each of the Dev and Eval test sets. These were extended for spoken language assessment (SLA) with a further (approximately) 150 submissions which had undergone phase 1 holistic proficiency scoring only. The submissions were selected so that the grade distribution was as even as possible but also reflects the nature of the learners using the platform. Table~\ref{tab:dev-dist} shows the grade distribution for the Dev set. As can be seen the majority are in the CEFR range B1-B2+. The Eval and Train sets have similar grade distributions. The scores in Table~\ref{tab:dev-dist} correspond to the marks awarded for a CEFR grade level on a part part basis. The overall score is the average of the scores across all the test parts so has a finer gradation. For this corpus the test is considered to consist of 4 parts. There was insufficient L1 and speaker data provided to incorporate this information into the data set selection. Note, this does mean that a speaker could appear in both the training and test sets. Table~\ref{tab:dev-eval-stats} shows the statistics for the Dev and Eval sets. 

\begin{table}[htbp]
    \centering
    \begin{tabular}{c|c|c}
    \toprule
       CEFR Grade & Score & \hspace{0.2cm}\% Data \hspace{0.2cm}\\
       \midrule
       A2  & 2.0 & 2.1 \\
       A2+  & 2.5 & 5.7 \\
       B1 & 3.0 & 18.3 \\
       B1+ & 3.5 & 25.3 \\
       B2 & 4.0 & 25.1 \\
       B2+ & 4.5 & 18.3 \\
       C1 & 5.0 & 5.0 \\
       C1+ & 5.5 & 0.2 \\
       \bottomrule
    \end{tabular}
    \caption{Distribution of holistic CEFR grades for Dev set. Eval and Train sets have similar distributions.}
    \label{tab:dev-dist}
\end{table}

\begin{table}[hbtp]
    \centering
    \begin{tabular}{c|cc|ccc|cc}
    \toprule
       Data Set & No. of & No. of & \multicolumn{3}{c|}{No. of Hours} & \multicolumn{2}{c}{No. of Words} \\
         & Submissions & Utterances & Trans & GEC & SLA & Transcript &\hspace{0.2cm}GEC\hspace{0.2cm} \\
         \midrule
    
         Dev & 438 & 5616 & 22.9 & 20.8 & 35.3 & 140k & 105k \\
         Eval & 442 & 5642 & 22.7 & 20.4 & 35.4 & 140k & 104k \\ 
         \bottomrule
    \end{tabular}
    \caption{Data set statistics for Dev and Eval set.}
    \label{tab:dev-eval-stats}
\end{table}

The Train set is a combination of data from full submissions and from individual parts, and is a mix of fully annotated and SLA annotated only data. The data was selected to be approximately equal across parts for the full data set, as shown in Table~\ref{tab:train-stats}. 

\begin{table}[hbtp]
    \centering
    \begin{tabular}{c|cc|ccc|cc}
    \toprule
       Part & No. of & No. of & \multicolumn{3}{c|}{No. of Hours} & \multicolumn{2}{c}{No. of Words} \\
         & Submissions & Utterances & Trans & GEC & SLA & Transcript &\hspace{0.2cm}GEC\hspace{0.2cm} \\ 
 \midrule
         1 & 3068 & 18072 & 7.3 & 3.9 & 70.3 & 47k & 22k \\
          3 & 3060 & 3060 & 5.8 & 2.9 & 47.0 & 37k & 14k \\ 
         4 & 3005 & 3005 & 5.8 & 1.5 & 45.5 &  36k & 7k \\
         5 & 3085 & 15353 & 9.4 & 4.8 & 81.4 & 60k & 24k \\
         \hline
       Total & 6640 & 39490 & 28.2 & 13.0 & 244.2 & 170k & 65k \\ 
       \bottomrule
    \end{tabular}
    \caption{Data set statistics for train set. 1742 submissions are complete for SLA scoring at an overall level.}
    \label{tab:train-stats}
\end{table}

\subsection{Corpus contents}
The corpus contents are listed in Table~\ref{tab:corpus-files}. Annotation markup and tags are described in Tables~\ref{tab:annotation-marks}, ~\ref{tab:ann-word-tags} and~\ref{tab:ann-phrase-tags}. All annotation was manually derived except for the time alignments provided. These were generated using the HTK HVite tool~\cite{htk341} adapted to do lattice-based forced alignment with an L2 English acoustic model.

\begin{table}[htbp]
    \centering
    \begin{tabular}{l|l}
    \toprule
    Item & Description \\
    \midrule
Audio files & 16kHz, single channel recordings. (flac)\\
\midrule
Audio file lists &  List of file IDs and their corresponding audio file. Correspond to data sets and subsets for specific tasks (tsv)\\
 \midrule
{Transcript annotations} & Marked up manual (disfluent) transcriptions with File IDs, question prompt and timing information. (json) \\
{GEC annotations} & As for manual transcript but the grammatical error corrected transcript. (json)\\
\midrule
    {SLA marks} & Holistic marks in the range 2-5.5 corresponding to A2-C1+ CEFR-like grades on a per-part and overall \\
    & submission level (tsv) \\
    \midrule
    STM transcriptions & Reference STMs for disfluent, fluent and GEC transcripts at the phrase level for use in NIST sclite scoring. \\
    & Contains automatically aligned phrase start and end times. \\
    & Category information included: audio quality (utterance level); grade (part); part number (part). (STM)\\
    \bottomrule
    \end{tabular}
    \caption{Material provided with the S\&I Corpus.}
    \label{tab:corpus-files}
\end{table}

\subsection{Use of the Corpus with LLMs}
We encourage the reader to consider the problem of ‘data leakage’ with regard to LLMs: whereby NLP/SLP
datasets are leaked into LLM training datasets via commercial APIs~\cite{balloccu-etal-2024-leak}. We therefore
ask that the Corpus is only passed to locally-stored LLMs, such as might be downloaded from the Hugging
Face Transformers library~\cite{wolf-etal-2020-transformers}, or to commercial LLMs in such a way that the data is not
retained for model training purposes.

\subsection{How to obtain the Corpus}
From December 2024-March 2025 the Speak \& Improve Corpus 2025 will be made available to participants in the \href{https://sites.google.com/view/slate-2025/home#h.p45mnlhypot5}{Speak \& Improve Challenge 2025} in connection with the ISCA SLaTE Workshop 2025, and will then be released for non-commercial academic research purposes. Access to the corpus may be obtained by visiting the \href{https://englishlanguageitutoring.com/datasets/speak-and-improve-corpus-2025}{ELiT website},
completing the form and agreeing to the licence when available. Future versions of the Corpus will be made available on
the same website. Queries about the corpus should be made to \href{mailto:support@speakandimprove.com}{support@speakandimprove.com}.

\section{Conclusions and Future Work}

In this paper we describe the Speak \& Improve Corpus 2025. This corpus of L2 learner English speaking provides a comprehensive resource for researchers interested in spoken language assessment and/or feedback.

The S\&I Corpus 2025 consists of audio and manual annotations of open speaking test submissions performed on the Speak \& Improve learning platform. We provide around 315 hours of speech data from L2 English speakers at A2 to C1 CEFR levels, split into train, dev and eval data sets. There are around 950 fully annotated test submissions and the equivalent of a further 2500+ test submissions with manual speaking proficiency assessment scores.

\section{Acknowledgements}
This paper reports on research supported by Cambridge University Press \& Assessment (CUP\&A), a department of The Chancellor, Masters, and Scholars of the University of Cambridge. 

Thanks to the team at \href{https://englishlanguageitutoring.com}{ELiT} who have created and run the Speak \& Improve learning platform, the ELiT annotation tool and the hosting and establishment of this corpus. In particular,  Paul Ricketts, Scott Thomas, Jon Phillips and Alex Watkinson.  
The ELiT humannotator team provided the annotations, with Gloria George and David Barnett providing phase 2 and 3 annotations.
Annabelle Pinnington and her team in Cambridge University Press \& Assessment's Multi-Level Testing team have provided all the question prompts for Speak \& Improve. Amanda Cowan and Jing Xu at Cambridge English have supported the development of this corpus and helped with all the paperwork. 

Thanks to the ALTA Spoken Language Processing Technology team at Cambridge University Engineering Department for helping to setup the Speak \& Improve Challenge 2025 and being beta-testers for the corpus, particularly Stefano Bann\`o, Penny Karanasou and Siyuan Tang.

The authors marked with$^\ast$ made equal contributions to this paper. 

\bibliographystyle{IEEEtran}
\bibliography{mybib,mybib2}

\end{document}